\documentclass[conference]{IEEEtran}

\usepackage{amsmath,amssymb,amsfonts,epsfig}
\usepackage{algorithm}
\usepackage[noend]{algpseudocode}
 \usepackage{mathtools}

\usepackage{array}
\usepackage{url}
\usepackage{graphicx}
\usepackage{caption}
\usepackage{subcaption}

\usepackage{color}
\usepackage{multirow}
\usepackage{array}
\ifCLASSINFOpdf
\else
\fi
\pdfoutput=1

\begin{document}

\title{A Unified Neural Network Approach for Estimating Travel Time and Distance for a Taxi Trip}

\author{\IEEEauthorblockN{Ishan Jindal}
\IEEEauthorblockA{Wayne State University\\
Detroit, MI\\
Email: ishan.jindal@wayne.edu}
\and
\IEEEauthorblockN{Tony (Zhiwei) Qin}
\IEEEauthorblockA{DiDi Research America\\
CA, USA\\
Email: qinzhiwei@didichuxing.com}
\and
\IEEEauthorblockN{Xuewen Chen}
\IEEEauthorblockA{DiDi Research\\
Beijing, China\\
Email: chenxuewen@didichuxing.com}
\and
\IEEEauthorblockN{Matthew Nokleby}
\IEEEauthorblockA{Wayne State University\\
Detroit, MI\\
Email: matthew.nokleby@wayne.edu}
\and
\IEEEauthorblockN{Jieping Ye}
\IEEEauthorblockA{DiDi Research\\
Beijing, China\\
Email: yejieping@didichuxing.com}}

\maketitle







\begin{abstract} \small\baselineskip=9pt In building intelligent transportation systems such as taxi or rideshare services, accurate prediction of travel time and distance is crucial for customer experience and resource management. Using the NYC taxi dataset, which contains taxi trips data collected from GPS-enabled taxis \cite{dataURL}, this paper investigates the use of deep neural networks to jointly predict taxi trip time and distance. We propose a model, called ST-NN (Spatio-Temporal Neural Network), which first predicts the travel distance between an origin and a destination GPS coordinate, then combines this prediction with the time of day to predict the travel time. The beauty of ST-NN is that it uses only the raw trips data without requiring further feature engineering 
and provides a joint estimate of travel time and distance. We compare the performance of ST-NN to that of state-of-the-art travel time estimation methods, and we observe that the proposed approach generalizes better than state-of-the-art methods. We show that ST-NN approach significantly reduces the mean absolute error for both predicted travel time and distance, about 17\% for travel time prediction. We also observe that the proposed approach is more robust to outliers present in the dataset by testing the performance of ST-NN on the datasets with and without outliers.
\end{abstract}
\section{Introduction}
Today, major cities in the world are expanding at a very fast pace. For these expanding cities, one of the potential problems is to efficiently utilize the existing road networks to reduce the potential traffic congestions. Therefore, the intelligent transportation systems are build such as advanced traveler information systems (ATIS) to minimize the traffic congestions by assisting the travelers in moving from one location to another. In ATIS, advance sensing technologies are used to acquire real time data either from in-road sensors such as loop detectors or from the mobile sensors such as GPS coordinates from moving vehicles. One such dataset collected from the mobile sensors is made available by the New York City Taxi \& Limousine Commission under the Freedom of Information Law (FOIL) \cite{dataURL} containing millions of taxi trips information. For each travel trip, this dataset provides information about the origin and destination GPS coordinates of the trip, travel time and travel distance of the trip, pickup date and time of the start and end of the trip and total fare. ATIS analyzes the acquired data and presents the relevant information to the user in the form of optimal routes, road conditions, the locations of incidents, travel time and distance estimation etc. \cite{schweiger2011use}.

In ATIS, estimated travel time and distance are very informative for travelers. This helps the traveler to plan their schedules in advance by using the potential traffic congestion information. Also, accurate measurement of travel time and distance helps in building intelligent transportation systems such as for developing the efficient navigation systems, for better route planning and for identifying key bottlenecks in traffic networks. The travel time and distance prediction depends heavily on the observable daily and weekly traffic patterns and also on the time-varying features such as weather conditions and traffic incidents. For instance, bad weather or an accident on road slows down the speed of the vehicles and cause lengthy travel time.

Most of the studies, in literature, for travel time estimation are focused on predicting the travel time for a sequence of locations, i.e  for a fixed route and commonly used techniques include (1) estimating travel time using historical data of travel trips; (2) using real time road speed information \cite{narayanan2015travel}\cite{zhang2011data}. The two common approaches for a route travel time estimation includes \emph{segment-based methods} and \emph{path-based methods}.

A simple approach for travel time estimation is the segment-based approach, in this approach the travel time is estimated on links (straight subsections of a travel path with no intersections ) first and then add them up to estimate the overall travel time. The link travel time is generally calculated by using loop detector data and floating car data \cite{kesting2013traffic} \cite{work2008ensemble} \cite{oh2002section} \cite{zhan2013urban}. Loop detectors sense the vehicle passing above the sensor and provides the continuous speed of the vehicle. This continuous speed information can then be used to calculate the travel time on that segment of road \cite{jia2001pems}. In addition to the loop detector, segment-based methods also use floating car data for travel time estimation \cite{de2008traffic}. Where, in floating car data, GPS enabled cars are used to collect timestamped GPS coordinates. The available dataset, in ST-NN, can think of as the special case of floating car data, where only the origin and destination GPS coordinates are recorded.

One of the major drawbacks of the segment-based method is that it can not capture the waiting times of a vehicle waiting at the traffic lights, which is a very important factor for estimating the accurate travel time. Therefore, some methods are developed which considers the waiting time at the intersections as well for travel time estimation \cite{li2015inferring} \cite{hofleitner2012learning}. In path-based methods, sub-paths (links + waiting time at intersections) are concatenated to predict the most accurate travel time \cite{hofleitner2012learning}. Our method is the special case of the path-based method, were sub-path is the entire path from origin to destination containing information about the waiting times at all the intersections. In addition to these methods, \cite{morgul2013commercial} propose a neighbor-based method for travel time estimation by averaging the travel time for all the samples in training data having the same origin, destination, and time-of-day.

In this paper, our focus is to jointly predict the travel time and distance from an origin to a destination as a function of the time-of-day using the historical NYC travel trips data. Since the available NYC taxi trip dataset does not contain GPS coordinates of the full trajectory of the trip, we treat it as a full path travel time estimation problem. One alternative solution for travel time estimation, can first find the specific trajectory path (route) between origin and destination and then estimate the travel time for that route \cite{gonzalez2007adaptive} \cite{yuan2010t}. Although, obtaining the travel route information is important, but we can think of a certain real scenario where route information is not as much important as travel time. For example, the travel route is of much less concern than the travel time to a non-driving taxi passenger. In \cite{zhan2013urban}, the historical taxi trip data is used for estimating the travel time by deriving the expected path travel time. It first selects all the probable travel path between an origin and a destination and takes the summation of each of the path travel time weighted by the probability of taking that particular path. 

In \cite{wu2004travel}, a Support Vector Regression (SVR) model is introduced for travel time estimation. The authors showed very promising results on a small highway dataset. Unlike highways, travel time variability is very high in urban cities because of traffic lights at each intersections \cite{yazici2014highway}, this makes it more challenging to predict the travel time in the cities. In this paper, we focus on the travel time and distance estimation between two locations in NYC. Since the publicly available taxi trips dataset contains information about millions of taxi travel trips and
influenced by the exceptional performance of deep neural networks \cite{schmidhuber2015deep}, given the tons of training data, we developed a unified deep neural network learning model that jointly learns the travel time and distance between an origin and a destination.

To the best of our knowledge, we are the first to estimate the travel distance directly from the GPS coordinates of origin and destination locations in the city, without building any route or map between the locations. 

In the subsequent sections, we first define the travel time estimation problem for origin-destination pair in Section \ref{ProbDef}, we briefly describe the multi-layer perceptron (MLP) and explain the ST-NN approach in Section \ref{Unified}. Then, we evaluate the performance of ST-NN approach in Section \ref{Experiments} and finally, we conclude this paper in Section \ref{Conclusion}.

\section{Problem Definition}
\label{ProbDef}
In this section, we explain the taxi travel time estimation problem in detail. Travel time is the time taken by a vehicle, moving from one location to another including the effect of temporal conditions. Similarly, travel distance is the distance transversed by a vehicle between two locations. In simple words, one can think of this problem as to estimate the travel distance and time between an origin (o) and a destination (d) at a particular time (t) time-of-day.

First, we define a taxi trip $p_i$, as a 5-tuple $(o_i, d_i, t_i, D_i, T_i)$, starting from the origin $o_i$ at time-of-day $t_i$ heading to the destination $d_i$, where $D_i$, is the travel distance and $T_i$ is the travel time from origin to destination. Both the origin and destination are 2-tuple GPS coordinates, that is $o_i = (\mathrm{Lat}_i, \mathrm{Lon}_i)$ and $d_i = (\mathrm{Lat}_i, \mathrm{Lon}_i)$, and time-of-day ($t_i$) is in seconds. An intuitive reason to include time-of-day $t_i$ as a part of taxi trip is that of different traffic conditions at the different time. For example, one can encounter heavy traffic at peak hours than off-peak hours also the traffic patterns on weekdays is different from weekends. Similar to \cite{wang2016simple}, we assume that the intermediate location or travel trajectory is not known, only the end locations are available. For reference, the largest publicly available NYC taxi trip dataset \cite{dataURL} contains only the end locations. We define a query $q_i$ as a pair $($\emph{origin, destination, time-of-day}$)_i $  input to the system and corresponding pair $($\emph{travel time, travel distance}$)_i$ as an output. Therefore, for the network, the only input query is  $(o_i, d_i, t_i)$, and the network estimates $(D_i, T_i)$. 

Given the historical database of $N$ taxi trips $\mathcal{X} = \{p_i\}_{i=1}^N$, our goal is to estimate the travel distance and time, $(D_q, T_q)$ for a query $q = (o_q, d_q, t_q)$,
\subsection{Data Mapping}
\label{subsetion:mapping}
\begin{figure}[h]
\centering
\includegraphics[width=\columnwidth]{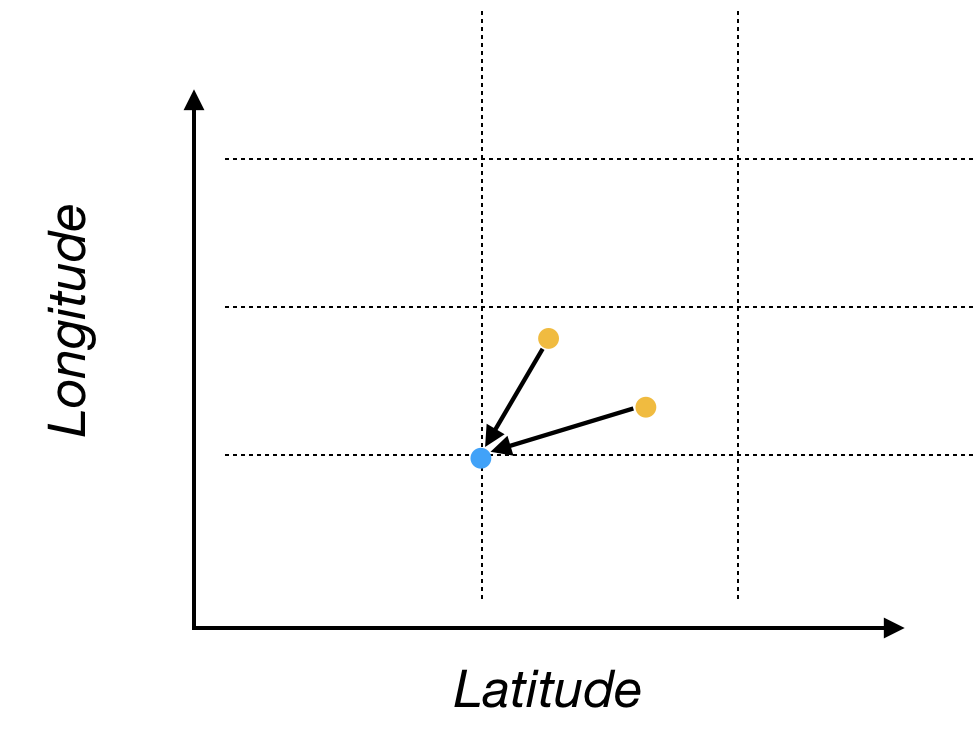}
\caption{GPS position binning}
\label{fig::LocBin}
\end{figure}

Geo-Coordinates are continuous variables and in the urban cities like NYC, because of tall buildings and dense areas, it is quite possible to get the erroneous GPS coordinates while reporting the data.  Other sources of erroneous recording of GPS coordinates involves atmospheric effects, multi-path effects and clock errors. For more information, we refer the reader to \cite{grewal2007global}. Therefore, to combat the uncertainties in GPS recording, a data pre processing step is needed to process the raw GPS data in order to get rid of erroneous GPS coordinates. Hence, we discretized the GPS coordinate into 2-D square cells, let's say of  $200mt.$ longitude and  $200mt.$ latitude. All the GPS coordinates of a square cell are represented by the lower left corner of that square cell as shown in Fig. \ref{fig::LocBin}. 

\begin{figure}[h]
\centering
\includegraphics[width=\columnwidth]{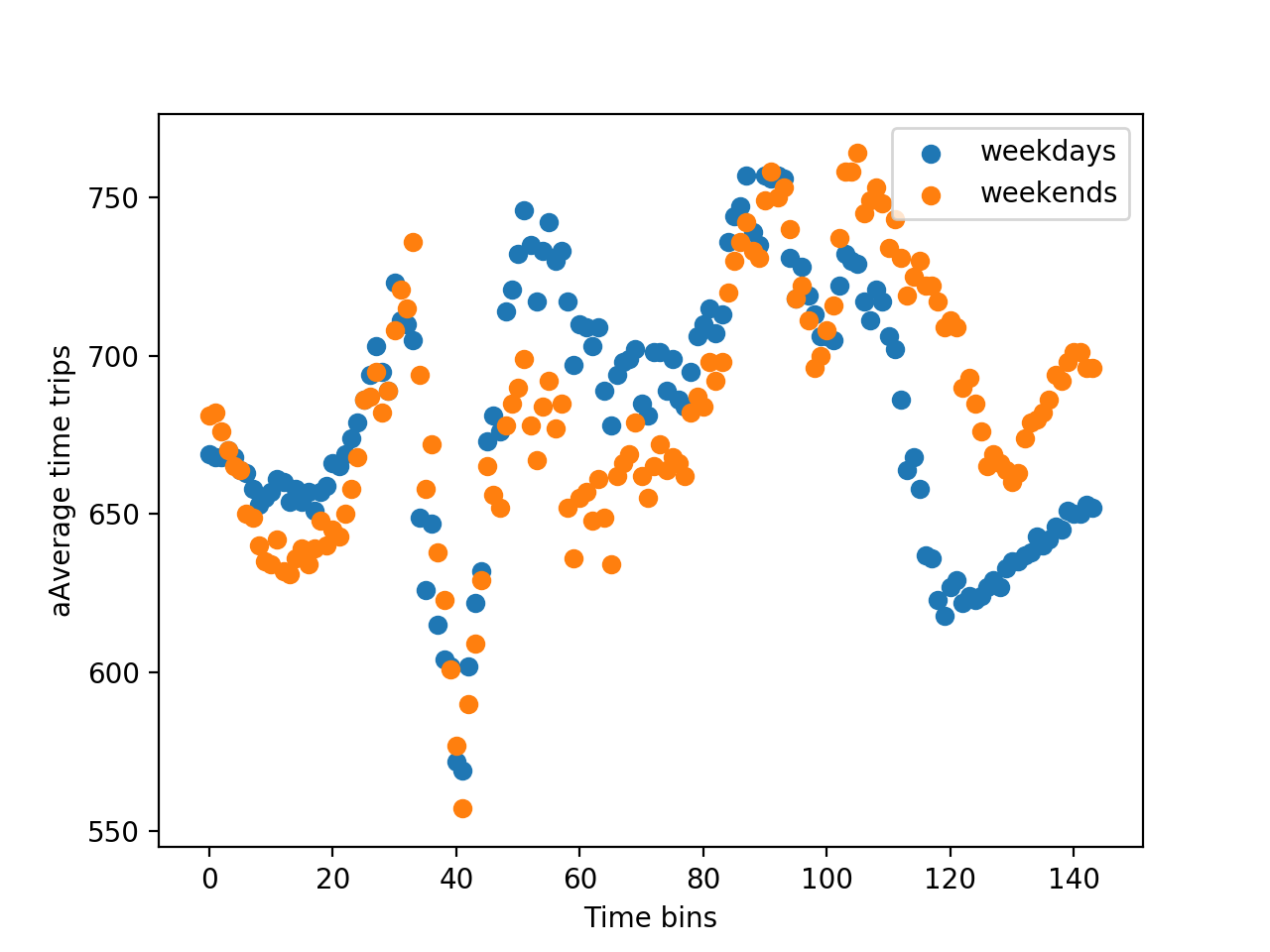}
\caption{Average taxi travel time per time cell on weekday and weekend}
\label{fig::TimDiff}
\end{figure}

Similar to location mapping, we also discretized the time-of-day as a 1-D time cell. From the NYC dataset, we observe that the average travel time of a taxi for weekday per time cell differs from the weekend as shown in Fig. \ref{fig::TimDiff}. 

\begin{figure}[h]
\centering
\includegraphics[width=\columnwidth]{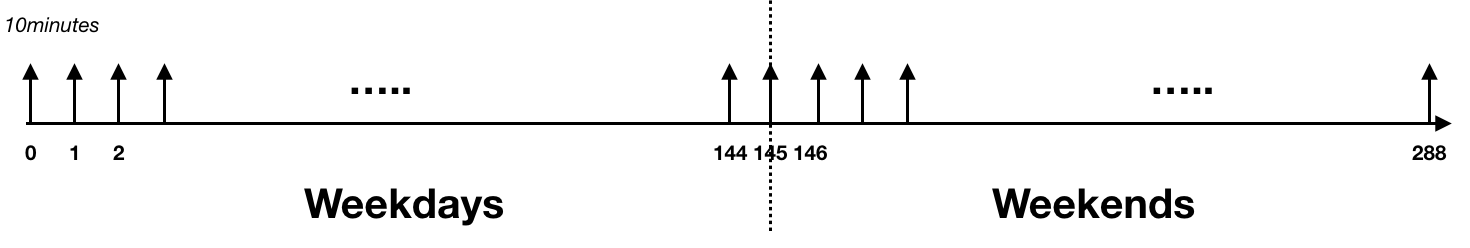}
\caption{Time binning}
\label{fig::TimBin}
\end{figure}

Therefore, we differentiate the time-of-day of weekdays from weekends. The time-of-day of the weekend is incremented by $3600*24$ sec. of time-of-day of the weekday,  as shown in Fig. \ref{fig::TimBin}, for a time cell of  $10$ Min. we obtain a total $288$ time cells. 

\section{Proposed Approach}
\label{Unified}

A simple possible solution can be to construct a look-up table containing average travel time and distance information for all possible queries. But this solution has several drawbacks. First, since the space formed by Cartesian product $o \times d \times t$ is very large, thus forming such a big look-up table requires a huge amount of memory to store. Second, most of the taxi application works in real time, that is updating the rider about remaining travel time. Therefore, frequently querying from such a big look-up table is time consuming and can not operate in real time. Finally, given the very sparse historical taxi trip data, it is not possible to have a query output for complete Cartesian product $o \times d \times t$ queries. Therefore, a regression based alternative approach is required.

\subsection{Background}

Deep neural networks are known for solving very difficult computational tasks like object recognition \cite{lecun1995learning} \cite{cirecsan2012deep}, regression \cite{west2000neural} and other predictive modeling tasks. They do so, because of their high ability to learn feature representations from the data \cite{hinton2006reducing} and best map the input features to the output variables. Also, neural networks are capable of learning any mapping from input features to output and can approximate any non-linear function \cite{hornik1989multilayer}. 

\begin{figure}[h]
\centering
\includegraphics[width=\columnwidth]{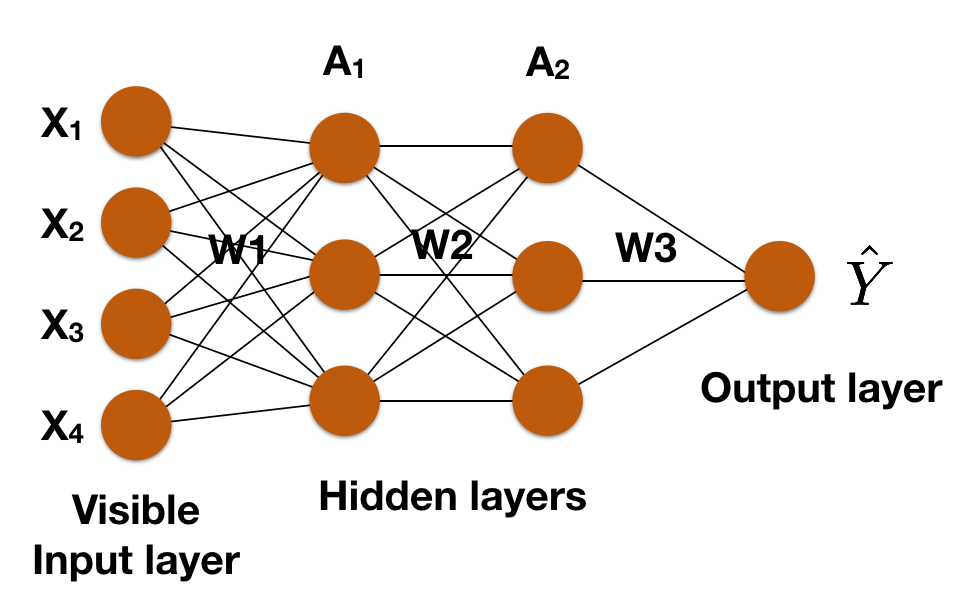}
\caption{Multi-Layer Perceptron}
\label{fig::MLP}
\end{figure}

In an artificial neural network, neurons serve as the basic building block of the networks. A neuron receives an input signal, process it using a logistic computation function and transmit an output signal depending on the computation outcome \cite{haykin2009neural}. When these neurons are arranged into networks of neurons termed as the artificial neural network. Each column of neurons in the network is called layer and a network can have multiple layers with multiple neurons each layer. Network with a single neuron is called perceptron and network with multiple layers of neurons is called multi-layer perceptron (MLP). A two hidden layer MLP is shown in Fig. \ref{fig::MLP}, where the input layer is the inputs to the network. The input layer is also called the visible layer because this the only exposed part of the network. Hidden layers derive features from the input layer at different scales or resolutions and form high-level features and output a value or a vector of values at the output layer depending on the type of the (regression, classification) problem.
\begin{figure*}[t!]
\centering
\includegraphics[width=0.75\textwidth]{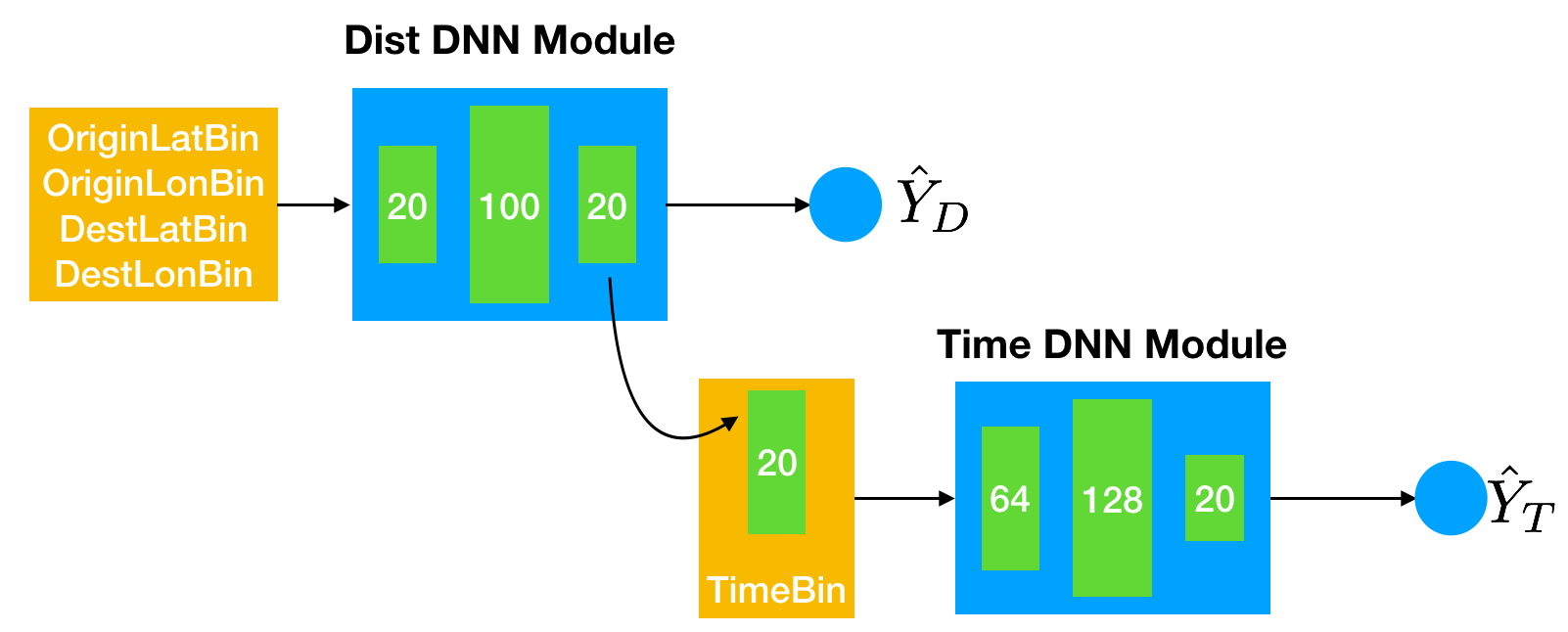}
\caption{Unified Neural Network Architecture for Joint Estimation of Travel Time and Distance}
\label{fig::ProbSchem}
\end{figure*}
At each hidden layer, network computes the features as:
\begin{align*}
A_1 &= f(W1*X)\\
A_2 &= f(W1*A_1)\\
Y &= f(W3*A_2)
\end{align*} 
Where $f$ is the activation function which takes the linear combination of weights and outputs at the previous layer and outputs a value and $*$ denotes the simple matrix multiplication. The activation function $f$ can be identical for all the hidden layers or can be different. $A_1$, $A_2$ and $\hat{Y}$ are the successive outputs of the first hidden layer, second hidden layer, and the final output layer. 

For a given row of data $X$ as an input to network and expected output $Y$, the network processes the input and obtains $A_1$, $A_2$ and finally obtain the predicted output $\hat{Y}$. This is called a forward pass. Then the predicted output is compared with the expected output $Y$ to compute an error using a loss function. The loss function measures our unhappiness with the outcome of the network. For example, in a regression problem, the mean square loss between predicted and expected output can be computed as:
\begin{equation}
L(Y, \hat{Y}) = \frac{1}{2N}\sum_{i=1}^N (Y^i-\hat{Y}^i)^2
\label{eq:loss}
\end{equation}
Where, $N$ is the number of training data samples and $Y^i$ represents the expected output of ith training sample. The empirical error computed according to \eqref{eq:loss} is then propagated back through the network using a standard backpropagation \cite{hecht1988theory} algorithm and updates the weights $W1, W2, W3$ for each layer according to a stochastic gradient descent algorithm, one layer at a time. This is called a backward pass. This process of a forward pass and a backward pass is repeated for all the data sample in training data and one pass over the entire training dataset is called an epoch. A network can be trained to minimize the loss for a large number of epochs.

All the hyper-parameters such as the number of layers in a network, the number of neurons per layer, activation of neurons, the loss function can be tuned by using multiple rounds of cross-validation. 

\subsection{ST-NN}
\label{sub:unified}
In Fig. \ref{fig::ProbSchem}, we describe the ST-NN architecture. In this architecture, we define two different deep neural network (DNN) module both for travel distance and time estimation as ``Dist DNN Module" and ``Time DNN Module'', respectively. First, we describe the inputs to both the modules. The input to dist DNN module is only the origin $o_i$ and destination $d_i$ binned GPS coordinates. This module is not exposed to time-of-day $t_i$ information because the time-of-day information is irrelevant to the travel distance estimation and might misguide the network. For any taxi service, because of usual reasons, always routes a driver on to a path that has the shortest length. As the route planning is not a part of this work, we assume that the all the taxis in the available taxi trip dataset have chosen the shortest path for a trip between origin and destination irrespective of time-of-day. Therefore, the input dimension to dist DNN module is 4-D, that is OriginLatBin, OriginLonBin, DestLatBin and DestLonBin. The input to time DNN module is the activations of last hidden layer of the dist DNN module along with the time-of-day information. Since, time-of-day is a very crucial parameter for estimating the travel time, as time-of-day carries daily and weekly traffic patterns and all the dynamic traffic condition information. 

Both the dist DNN module and time DNN module are three-layer MLP with different numbers of neurons per layer. We cross-validated the parameters and find the ones with the best performance. The best performance configuration of the number of layers and number of neurons per layer for both the module is shown in Fig. \ref{fig::ProbSchem} where, $\hat{Y}_D$ and $\hat{Y}_T$ are the predicted distance and time from dist DNN module and time DNN module, respectively. The ST-NN architecture is then trained via stochastic gradient descent jointly for both travel distance and time according to the loss function:
\begin{equation}
L(Y_D, Y_T,\hat{Y}_D,\hat{Y}_T) = L(Y_T, \hat{Y}_T) + L(Y_D ,\hat{Y}_D)
\end{equation}
From \eqref{eq:loss}, we write the final loss function as:
\begin{multline}
L(Y_D, Y_T,\hat{Y}_D,\hat{Y}_T) = \frac{1}{2N}\sum_{i=1}^N (Y_T^i-\hat{Y}_T^i)^2 + \\ \frac{1}{2N}\sum_{i=1}^N (Y_D^i-\hat{Y}_D^i)^2
\end{multline}	

We observe, in Section \ref{Experiments}, that the joint learning of travel distance and time as in Fig. \ref{fig::ProbSchem} improves the travel time estimation over the baseline methods, described in next section.
\subsection{DistNN}
\label{sub:dist}
\begin{figure}[h]
\centering
\includegraphics[width=\columnwidth]{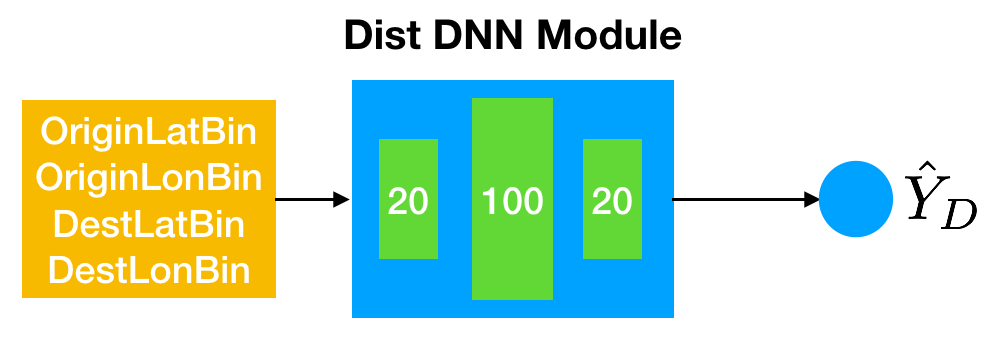}
\caption{MLP architecture for travel distance estimation}
\label{fig::Dist_Module}
\end{figure}

For comparing the performance of ST-NN with the standalone modules, we also define a three layer MLP regression network for estimating the travel distance from the origin $o_i$ and destination $d_i$ binned GPS coordinates as shown in Fig. \ref{fig::Dist_Module}.

Here, we use the same network parameters of dist DNN module as in Fig. \ref{fig::ProbSchem} and train the network via stochastic gradient descent with the loss function:
\begin{align}
L(Y_D,\hat{Y}_D) &= L(Y_D ,\hat{Y}_D)\\
&=  \frac{1}{2N}\sum_{i=1}^N (Y_D^i-\hat{Y}_D^i)^2
\end{align}

\subsection{TimeNN}
\label{sub:time}
\begin{figure}[h]
\centering
\includegraphics[width=\columnwidth]{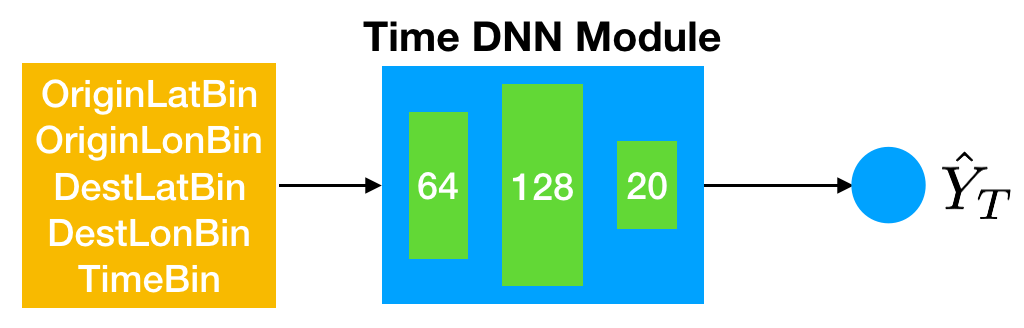}
\caption{MLP architecture for travel time estimation}
\label{fig::Time_Module}
\end{figure}

Similar to standalone dist DNN module, we also show the performance of standalone time DNN module for travel time estimation as shown in Fig. \ref{fig::Time_Module}. We use the same network parameters for time DNN module as in Fig. \ref{fig::ProbSchem}. The input to this network are origin $o_i$ and destination $d_i$ binned GPS coordinates along with time-of-day information. 

Finally, the network is trained via stochastic gradient descent with the loss function:
\begin{align}
L(Y_T,\hat{Y}_T) &= L(Y_T, \hat{Y}_T)\\
&= \frac{1}{2N}\sum_{i=1}^N (Y_T^i-\hat{Y}_T^i)^2
\end{align}

\section{Experiments and Results}
\label{Experiments}

In this section, at first we start by describing the NYC taxi dataset used for evaluating the performance of ST-NN and then we describe the evaluation measure used in paper. Afterwards, we define performance measures and results.
\subsection{NYC Dataset}
\begin{figure}[h]
    \centering
    \begin{subfigure}[t]{0.5\columnwidth}
        \centering
        \includegraphics[width = \columnwidth]{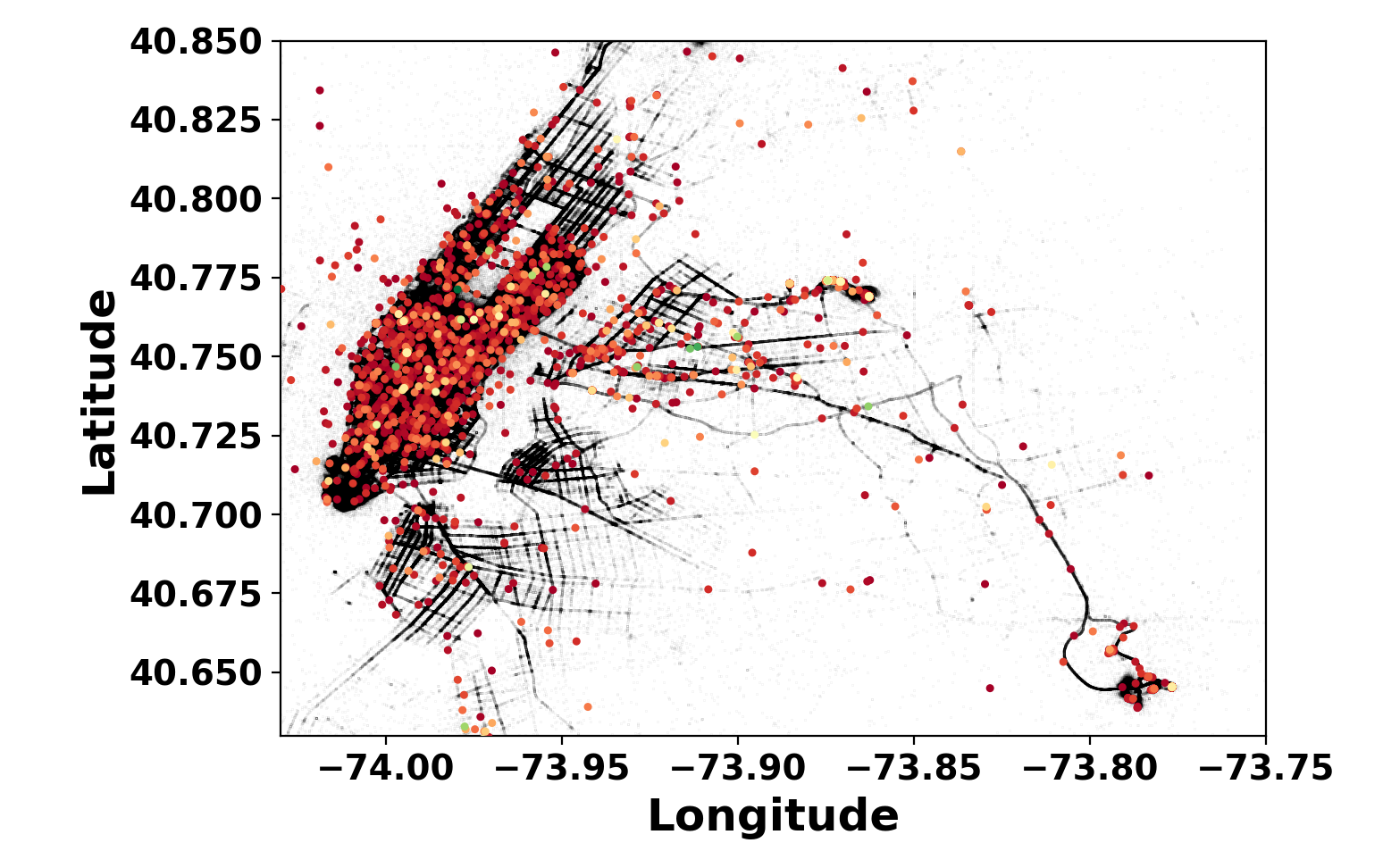}
        \caption{Taxi pickup Distribution}
        \label{fig::PickDist}
    \end{subfigure}%
    ~ 
    \begin{subfigure}[t]{0.5\columnwidth}
        \centering
        \includegraphics[width = \columnwidth]{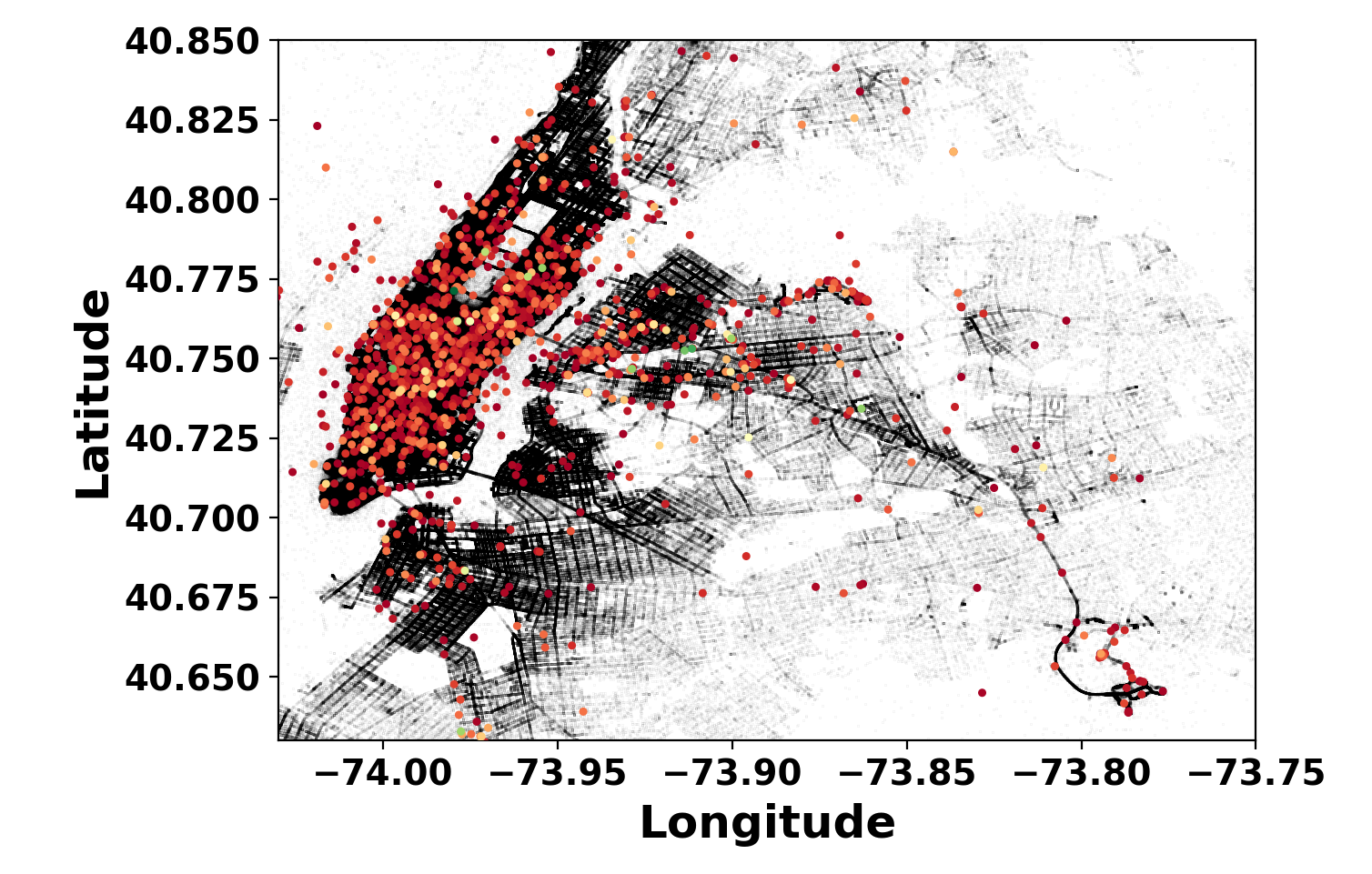}
        \caption{Taxi dropoff Distribution}
        \label{fig::DropDist}
    \end{subfigure}
    \caption{NYC GPS Coordinates Distribution}
    \label{GPS}
\end{figure}

A publicly available gigantic taxi trip dataset, recorded $173$M taxi trips for the New York City during the year 2013 \cite{dataURL}.  This dataset describes every single trip by 21 different variables containing the pickup GPS coordinates, dropoff GPS coordinates, date and time of pickup and dropoff, total travel time in seconds, total travel distance in miles, the number of passengers, fare amount, tax amount, driver's license, rate code etc. Fig. \ref{GPS} outline the provided GPS coordinates where Fig. \ref{fig::PickDist} and  \ref{fig::DropDist} show the density of pickup and dropoff GPS coordinates, respectively. 

\begin{figure}[h]
    \centering
    \begin{subfigure}[t]{0.5\columnwidth}
        \centering
        \includegraphics[width = \columnwidth]{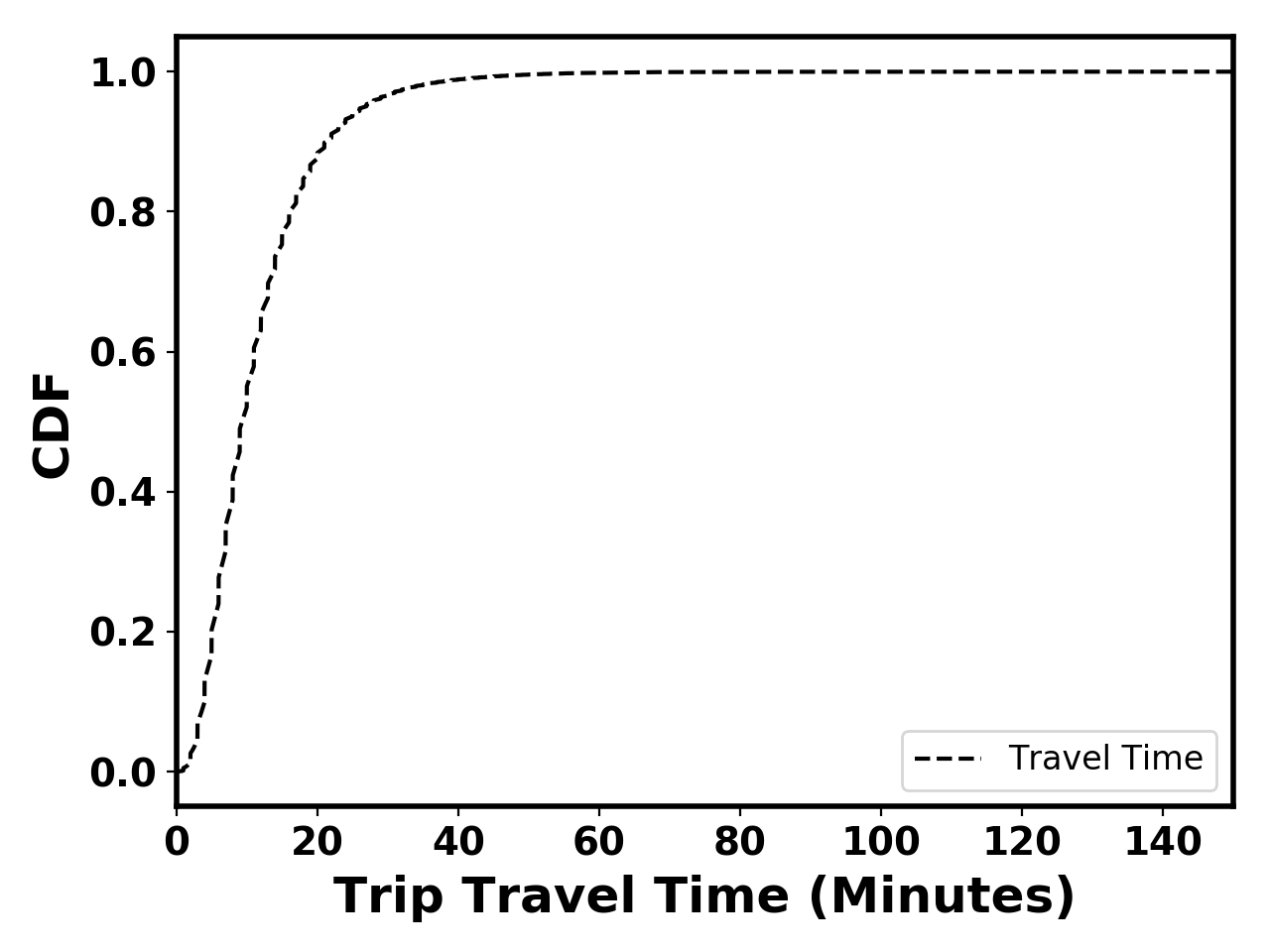}
        \caption{Travel Time CDF}
        \label{fig::CDFTime}
    \end{subfigure}%
    ~ 
    \begin{subfigure}[t]{0.5\columnwidth}
        \centering
        \includegraphics[width = \columnwidth]{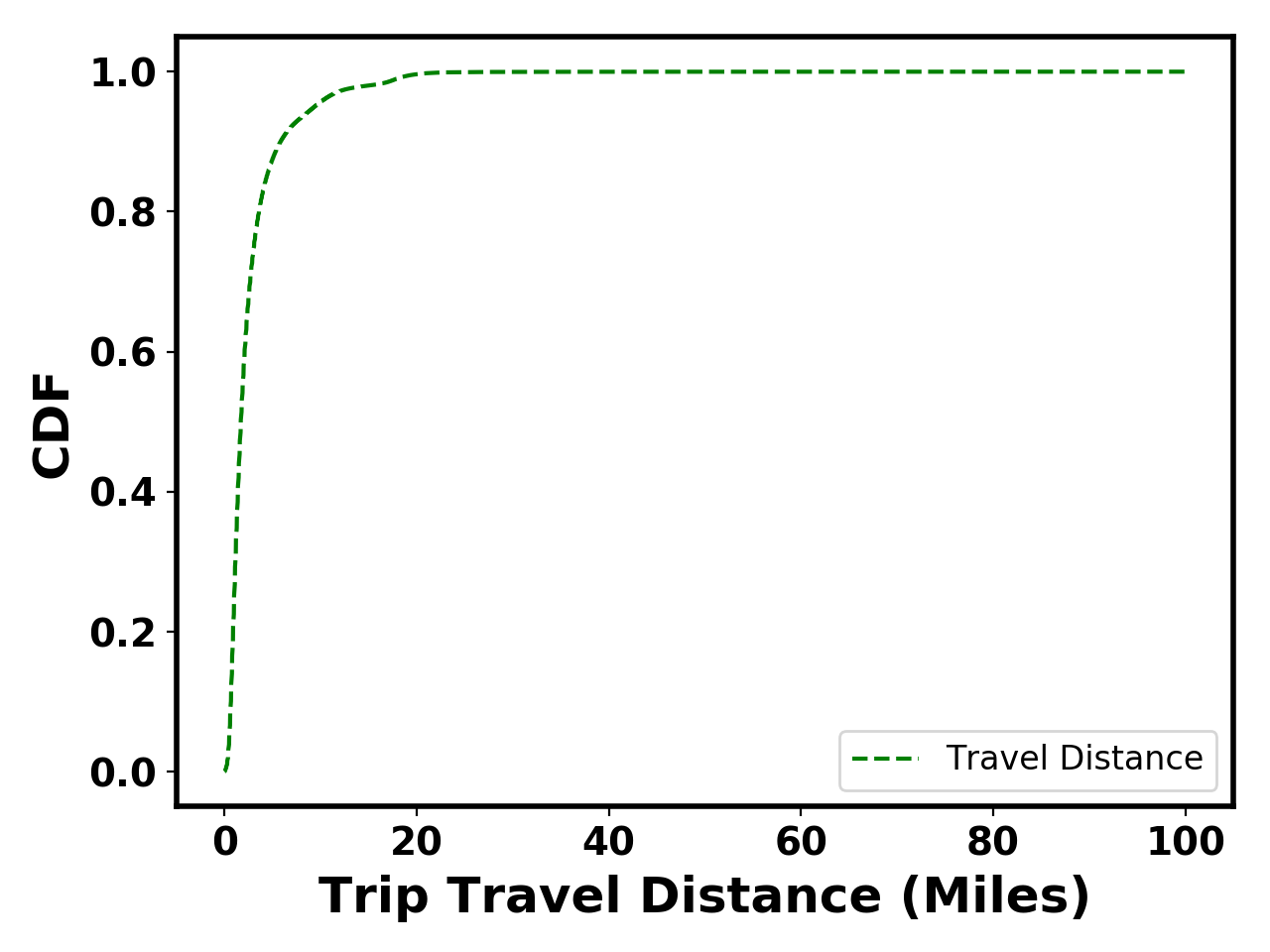}
        \caption{Travel Distance CDF}
        \label{fig::CDFDist}
    \end{subfigure}
    \caption{NYC Taxi Trips Statistics}
    \label{lab:cdf}
\end{figure}

We also provide some statistics of the data in Fig. \ref{lab:cdf}. We show the empirical CDF plots for travel time and distance in Fig. \ref{fig::CDFTime} and \ref{fig::CDFDist}. From the dataset, we observe that about 80\% of the travel trips have travel time less than 20 minutes and about 60\% of the trips have travel distance less than 2 miles.


\subsection{Evaluation Methods} 
\label{sub:evaluation_methods}
Here, we describe the methods we compared with our proposed approach.\\
\begin{enumerate}
	\item Linear Regression for Time (\emph{LRT}): We implement a simple linear regression method for time estimation, modeling travel time as a function of origin and destination GPS coordinates defined by the 2-D square cell and the 1-D time cell number. 
	\item Linear Regression for Distance (\emph{LRD}): Similarly, We implement a simple linear regression method for distance estimation, modeling travel distance as a function of origin and destination GPS coordinates defined by the 2-D square cell.
	\item Time DNN module (\emph{TimeNN}): This is the method when only the time DNN module of the unified network is used to learn the travel time as described in Section \ref{sub:time}. Inputs to this module are the origin and destination GPS coordinates defined by the 2-D square cell and the 1-D time cell number.
	\item Distance DNN module (\emph{DistNN}): Similarly, This is the method when only the distance DNN module of the unified network is used to learn the travel distance as described in Section \ref{sub:dist}. Inputs to this module are only the origin and destination GPS coordinates defined by the 2-D square cell.
	\item Unified learning (\emph{ST-NN}): This is the proposed improved approach described in Section \ref{sub:unified}.
	\item (BTE) \cite{wang2016simple} : We also compare the performance of proposed approach with the best method introduced in \cite{wang2016simple}.
\end{enumerate}

\subsection{Outliers Rejection}
\label{subsec:outlier}
From the initial exploration of NYC taxi trip data we find that the dataset contains a number of anomalous taxi trips termed as outliers, for example having more than 7 passengers in a taxi and no passenger, missing pickup and dropoff GPS coordinates, travel time of zero seconds while the corresponding travel distance is non-zero, travel distance of zero miles while corresponding travel time is non-zero. 

These outliers can cause huge mistakes in our estimations so, we experimentally detected the anomalous trips and removed from the dataset. We also defined a GPS coordinate box for the NYC, obtained from \cite{NYCURL}, and select a subset of taxi trips within the borough of this GPS coordinate box in order to remove all the taxi trips having pickup or dropoff GPS coordinates lies outside the NYC.

\subsection{Performance evaluation}
\label{sec:Performance_evaluation}

First, we describe the performance measure used in this paper to evaluate the performance of ST-NN. We acquire the measures from \cite{wang2014travel} (1) to evaluate the travel time estimation and (2) to have a fair comparison with \cite{wang2016simple}, such as \emph{Mean Absolute Error (MAE)} and \emph{Mean Relative Error (MRE)}. Where MAE is defined as the mean of the absolute difference between the estimated travel time $f_i$ and the ground truth $y_i$:
\begin{equation}
	\mathrm{MAE} = \frac{\sum_{i=1}^N |y_i - f_i|}{N}
\end{equation}
and, MRE is defined as:
\begin{equation}
	\mathrm{MRE} = \frac{\sum_{i=1}^N |y_i - f_i|}{\sum_{i=1}^N y_i}
\end{equation}
We also define the \emph{Median Absolute Error (MedAE)} and \emph{Median Relative Error (MedRE)}, as the dataset contains anomalous taxi trip entries that is $$ \mathrm{MedAE} = \mathrm{median}( |y_i - f_i|), \mathrm{MedRE} = \mathrm{median}\left(\frac{|y_i - f_i|}{y_i}\right),$$ where $\mathrm{median} $ has its usual meaning. To measure how close the data are to the fitted hyper surface, we also use the coefficient of determination $R^2$ to evaluate the performance of proposed approach. $R^2$ coefficient ranges between $(-\infty , 1]$, where the -ve value indicates that the fitted hyper surface accounts none for the variation in data and $R^2 = 1$ indicates that all the data points perfectly fall on the fitted hyper surface. $R^2$ coefficient is defined as $$ R^2 = 1- \frac{\sum_{i}(y_i - f_i)^2}{\sum_{i} (y_i - \bar{y})^2}.$$ Where $\bar{y} = \frac{1}{N}\sum_{i=1}^N y_i$, is the mean of the observed data. 

\subsection{Results} 
\label{sec:results}
We evaluate the performance of ST-NN on the NYC travel trip dataset. We divide the entire dataset into training and test subsets in the ratio 80:20. All the results are reported on the test subset. All the parameters of ST-NN network architecture such as a number of layers per module and number of units per hidden layer is shown in the Fig. \ref{fig::ProbSchem}. We also use data mapping as described in Section \ref{subsetion:mapping}. For location mapping, we use $(200mt. \times 200mt.)$ 2-D square cell and for time mapping we use 10 minutes as 1-D time cell. To compare the performance of TimeNN and DistNN with the proposed approach we use the same module parameters as in time DNN module and distance DNN module respectively. All the parameters are kept fixed of ST-NN throughout all the experiments. 
\begin{table*}[]
	\centering{}
	\begin{tabular}{c|c|c|c|c|c}
		          & $R^2$ Coefficient & MAE & MRE & MedAE & MedRE \\ \hline
		LRT       &     -1.84                            &  724.14   & 1.01    &   638.52    &   1.10    \\ \hline
		TimeNN    &      0.713                          &   158.29  &   0.221  &    100.242   &   0.182    \\ \hline
		ST-NN &       0.75                          & 145.9    &  0.20   &    91.48   &     0.16 
	\end{tabular}
	\caption{Overall performance comparison of proposed approach with the other approaches for travel time estimation, when trained on entire NYC}
	\label{tabel:Time_comp}
\end{table*}

\begin{figure}[htbp]
\centering
\includegraphics[width=3.in]{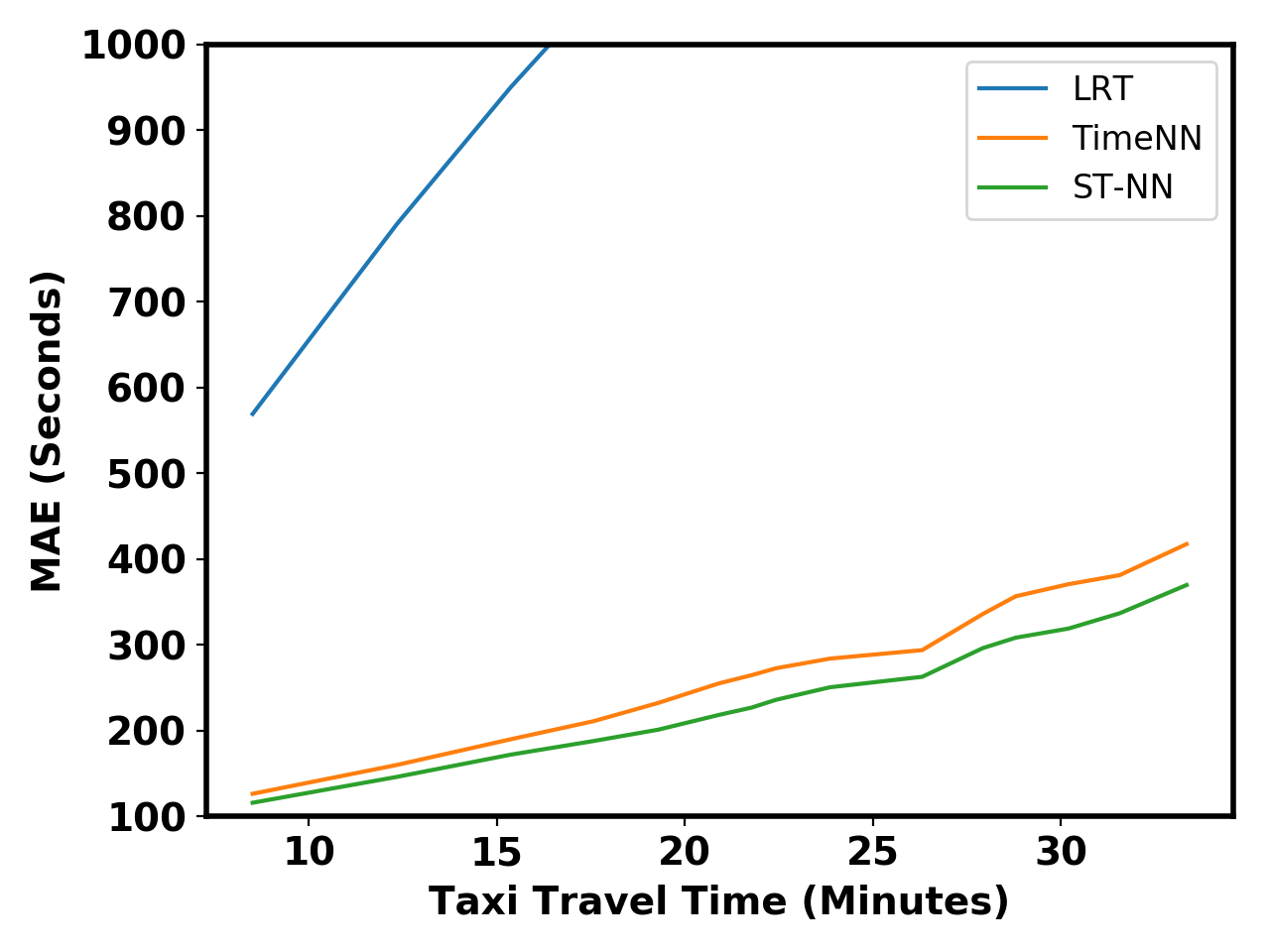}
\caption{Overall performance comparison as a function of taxi travel time}
\label{fig::TimeComp}
\end{figure}

Table \ref{tabel:Time_comp} and \ref{tabel:Dist_comp} compare the performance of proposed approach for travel time and distance estimation, respectively. From Table \ref{tabel:Time_comp}, we first observe that the proposed approach is far better than the simple linear regression method for travel time estimation. This is expected because the simple linear regression is a baseline method which does not consider the uncertain traffic conditions and simply tries to find the linear relationship between the raw origin-destination GPS coordinates and the travel time. Considering the temporal differences, TimeNN is the big shot and improves the travel time prediction a lot better just mapping the raw origin-destination GPS coordinates and time-of-day information to travel time. We observe the huge differences in all performance measures and about $78\%$ improvement in MAE. 

\begin{figure}[h]
\centering
\includegraphics[width=\columnwidth]{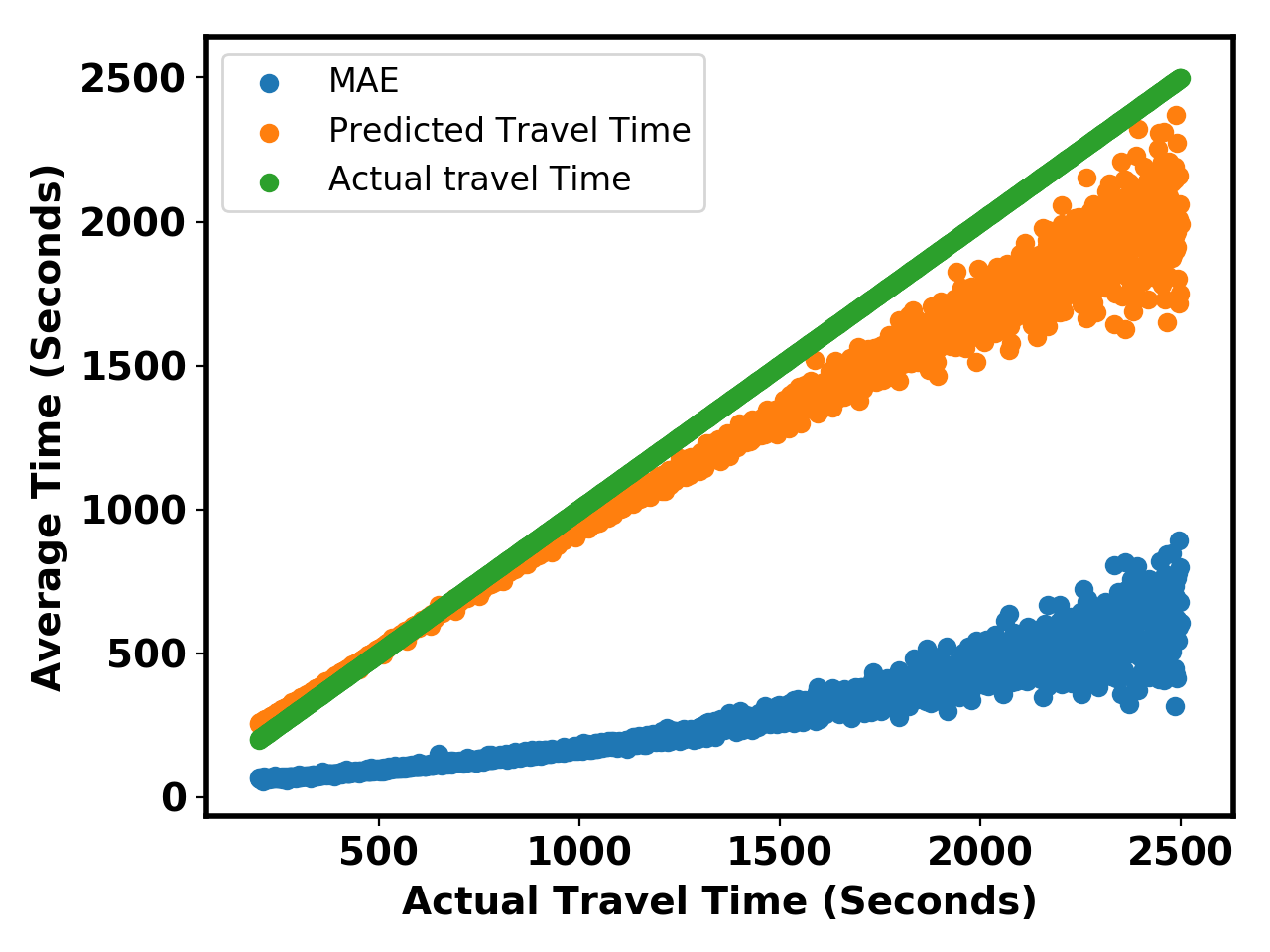}
\caption{ST-NN performance as a function of taxi travel time}
\label{fig::TraveMAE}
\end{figure}

By adding the encoded travel distance information (the ST-NN), further improves the performance for travel time estimation, that is MAE is improved by 13 seconds in comparison to TimeNN. For reference, this MAE is average over millions of taxi trips so, thus the difference of 13 seconds in MAE means a lot. To investigate further, we plot the MAE for all the approaches in Fig. \ref{fig::TimeComp} to know in which regimes the ST-NN is better than the TimeNN. It is clear from the curves that the slope of the orange curve is more than the green curve, that is as the taxi travels far a significant gap in the performance is noticed. We also plot the MAE and predicted travel time for ST-NN network as a function of taxi travel time in Fig. \ref{fig::TraveMAE}. As expected, for the shorter taxi trips ST-NN succeeds in predicting the actual travel time but for the longer travel trips, it encounters a larger MAE, around $8-10$ minutes. Since the short travel time trips are more prone to be affected by the temporal conditions, we can say that the proposed approach is all time better at capturing the dynamic conditions.

We also evaluate the performance of ST-NN for travel distance estimation. Similar to travel time estimation, from table \ref{tabel:Dist_comp}, we find that the LRD performs very poorly. It is because, in any urban city, it is very hard to find a straight route from the origin to destination. Therefore, finding a linear distance pattern is not an efficient approach, however, LRD always tries to strive for a linear pattern. We also observe a very small performance difference between the DistNN and the ST-NN, DistNN performed better than the ST-NN. To explain this performance gap, we say that the ST-NN compromises a little bit in performance for travel distance estimation to obtain high performance on travel time estimation.

\begin{table*}[h]
	\centering
	\begin{tabular}{c|c|c|c|c|c}
		          & $R^2$ Coefficient & MAE & MRE & MedAE & MedRE \\ \hline
		LRD       &         -0.397                        &  3.109   &  1.045   &   2.549    &  1.224     \\ \hline
		DistNN    &           0.95                      &   0.21  &  0.07   &    0.077   &   0.0418    \\ \hline
		ST-NN &           0.943                     &  0.27   &   0.09  &   0.112    &      0.06
	\end{tabular}
	\caption{Overall performance comparison of proposed approach with the other approaches for travel distance estimation, when trained on entire NYC}
	\label{tabel:Dist_comp}
\end{table*}

\begin{figure}[h]
\centering
\includegraphics[width=3.in]{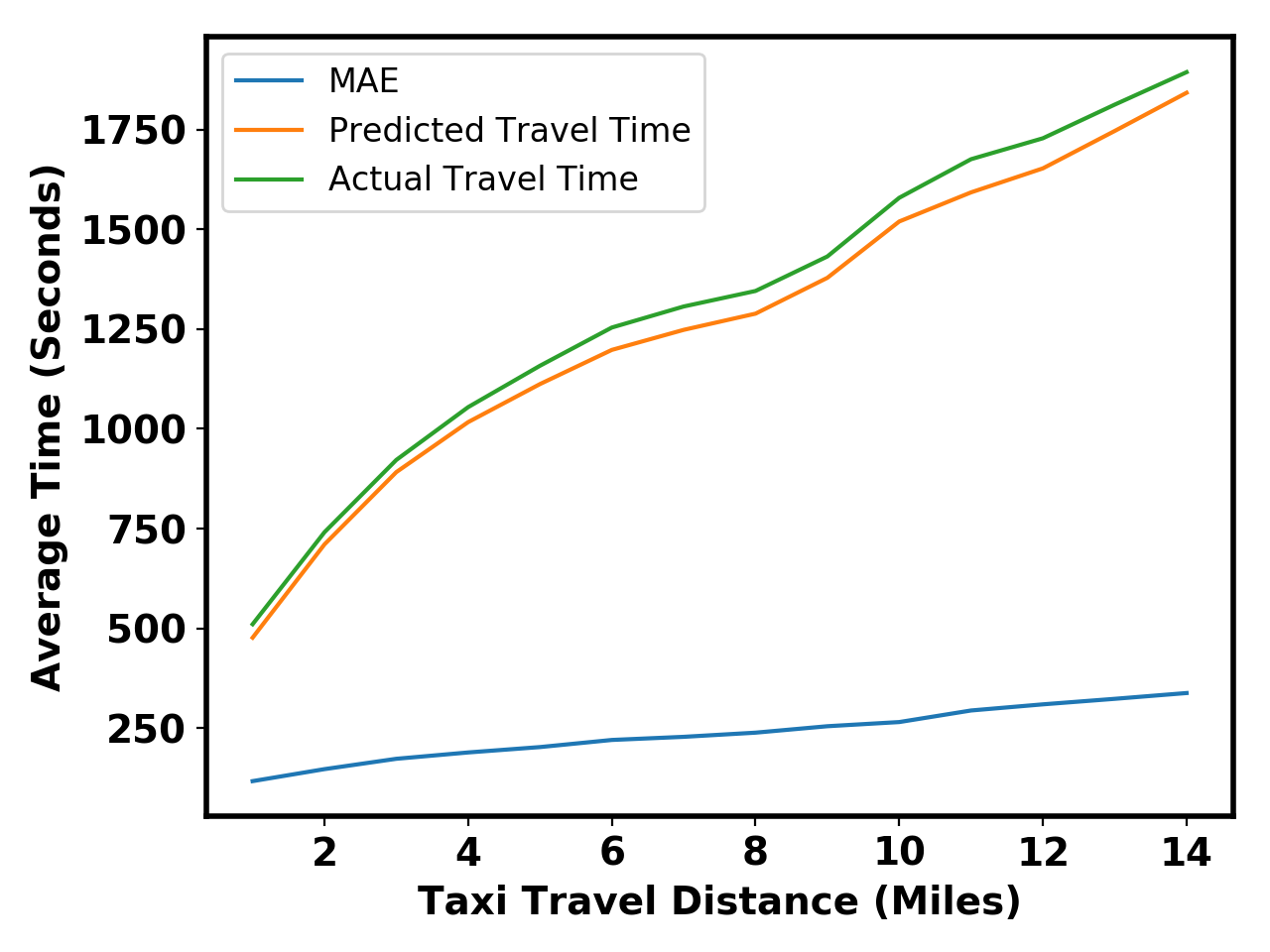}
\caption{ST-NN Performance as a function of taxi travel distance}
\label{fig::DistMAE}
\end{figure}

We also evaluate the performance of ST-NN for travel time estimation with respect to the trip distance in Fig. \ref{fig::DistMAE}. We observe the similar behavior as estimated travel time with respect to the actual travel time in Fig. \ref{fig::TraveMAE}, that is as the taxi travels far, MAE increases. This is because of the reason that a long distance trip always has long travel time. 

In this paper, we compare the performance of proposed approach with the best approach in \cite{wang2014travel} with the same parameter settings. As the authors in \cite{wang2014travel}, report result only within the borough of Manhattan, we also mask the training dataset confined only to Manhattan. We also use the same data mapping parameters: for location $(50mt. \times 50mt.)$ as 2-D square cell and for time 60 minutes as 1-D time cell. We summarize the performance of ST-NN with respect to BTE \cite{wang2014travel} in Table \ref{tabel:paper_Comp}. We observe a clear performance improvement of ST-NN for travel time estimation, in terms of MAE, by 17\%.

We can not provide the direct comparison for travel distance estimation because \cite{wang2014travel} only showed results on travel time estimation.

\begin{table}[h]
	\centering
	\begin{tabular}{c|c|c|c|c}
		            & MAE    & MRE    & MedAE  & MedRE  \\ \hline
		ST-NN   & 121.48 & 0.215  & 80.77  & 0.182  \\ \hline
		BTE \cite{wang2014travel} & 142.73 & 0.2273 & 98.046 & 0.1874
	\end{tabular}
	\caption{Performance of proposed approach compared to \cite{wang2014travel}}
	\label{tabel:paper_Comp}
\end{table}

We also study the impact of outliers on the performance of our approach and compare it with \cite{wang2014travel} for travel time estimation in Table \ref{label:w_o_comp}. In Section \ref{subsec:outlier}, we studied the types of outliers present in dataset and applies certain filters on the dataset such as filters using time and distance, GPS coordinates etc to remove the outliers. To analyze the robustness of ST-NN with respect to outliers, we train the ST-NN on the cleaned training data and test the network on uncleaned (with outliers) data. We found that when the outliers are prevalent in the data, our proposed approach not only outperform \cite{wang2014travel} but also appears to be more robust to outliers. We observe a difference of approx. 2 seconds in MAE for the proposed approach.

\begin{figure}[h]
\centering
\includegraphics[width=\columnwidth]{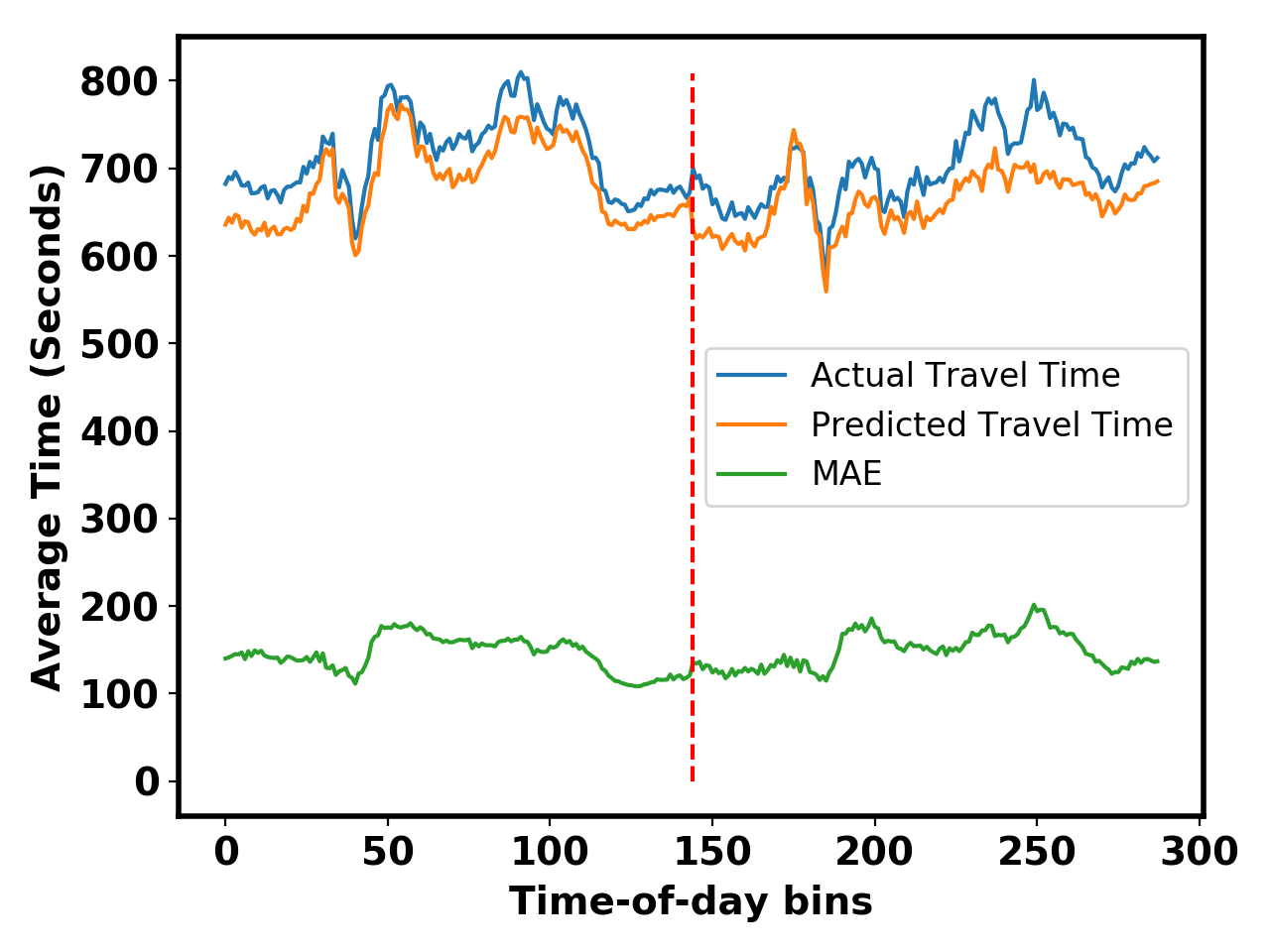}
\caption{ST-NN Performance: MAE vs. Time-of-day}
\label{fig::MaeVSTime}
\end{figure}

\begin{table}[h]
	\centering
	\begin{tabular}{c|c|c|c}
		\multicolumn{2}{|c|}{}                                                                    & MAE    & MRE    \\ \hline
		\multirow{2}{*}{\begin{tabular}[c]{@{}c@{}}With \\ Outliars\end{tabular}}   & ST-NN   & 123.13 & 0.2282 \\ \cline{2-4} 
		                                                                            & BTL \cite{wang2014travel} & 170.04 & 0.2547 \\ \hline
		\multirow{2}{*}{\begin{tabular}[c]{@{}c@{}}Without\\ outliars\end{tabular}} & ST-NN   & 121.48 & 0.2155 \\ \cline{2-4} 
		                                                                            & BTL \cite{wang2014travel}  & 142.73 & 0.2173 \\ 

	\end{tabular}
	\caption{Performance of the proposed approach with/without outlier}
	\label{label:w_o_comp}
\end{table}
We plot the MAE of ST-NN as a function of time-of-day in Fig. \ref{fig::MaeVSTime} to comparing the performance of ST-NN for weekday vs. weekends. From the plot, the time bins on the left of the dotted red line represent the weekday and on the right are the weekends. We find that for weekdays, the predicted travel time form ST-NN model successfully mimics the actual travel time, it is because that a large number of people commute to work and home during weekdays and forms a pattern which is learned by the neural network. But for weekends, during the later times in the day, when there are unusual traffic patterns, the ST-NN can not  capture the temporal patterns as efficiently as compared to weekdays.

\section{Conclusion}
\label{Conclusion}
In this paper, we propose a ST-NN deep neural network approach for jointly estimating the travel time and travel distance from the historical travel trip dataset. The ST-NN just uses the raw origin and destination GPS coordinates and time-of-day information without requiring any feature engineering to efficiently predict the travel time and distance. The ST-NN not only outperformed the state-of-the-art methods for travel time estimation but also maintains its performance in the presence of outliers. Finally, the ST-NN provides the results for both travel time and travel distance estimation.

\bibliographystyle{IEEEtran}
\bibliography{ref.bib}

\begin{thebibliography}{10}
\providecommand{\url}[1]{#1}
\csname url@samestyle\endcsname
\providecommand{\newblock}{\relax}
\providecommand{\bibinfo}[2]{#2}
\providecommand{\BIBentrySTDinterwordspacing}{\spaceskip=0pt\relax}
\providecommand{\BIBentryALTinterwordstretchfactor}{4}
\providecommand{\BIBentryALTinterwordspacing}{\spaceskip=\fontdimen2\font plus
\BIBentryALTinterwordstretchfactor\fontdimen3\font minus
  \fontdimen4\font\relax}
\providecommand{\BIBforeignlanguage}[2]{{%
\expandafter\ifx\csname l@#1\endcsname\relax
\typeout{** WARNING: IEEEtran.bst: No hyphenation pattern has been}%
\typeout{** loaded for the language `#1'. Using the pattern for}%
\typeout{** the default language instead.}%
\else
\language=\csname l@#1\endcsname
\fi
#2}}
\providecommand{\BIBdecl}{\relax}
\BIBdecl

\bibitem{dataURL}
C.~Whong, ``Foiling nyc boro taxi trip data,''
  \url{http://chriswhong.com/open-data/foiling-nycs-boro-taxi-trip-data/}.

\bibitem{schweiger2011use}
C.~L. Schweiger, \emph{Use and Deployment of Mobile Device Technology for
  Real-time Transit Information}.\hskip 1em plus 0.5em minus 0.4em\relax
  Transportation Research Board, 2011, vol.~91.

\bibitem{narayanan2015travel}
A.~Narayanan, N.~Mitrovic, M.~T. Asif, J.~Dauwels, and P.~Jaillet, ``Travel
  time estimation using speed predictions,'' in \emph{Intelligent
  Transportation Systems (ITSC), 2015 IEEE 18th International Conference
  on}.\hskip 1em plus 0.5em minus 0.4em\relax IEEE, 2015, pp. 2256--2261.

\bibitem{zhang2011data}
J.~Zhang, F.-Y. Wang, K.~Wang, W.-H. Lin, X.~Xu, and C.~Chen, ``Data-driven
  intelligent transportation systems: A survey,'' \emph{IEEE Transactions on
  Intelligent Transportation Systems}, vol.~12, no.~4, pp. 1624--1639, 2011.

\bibitem{kesting2013traffic}
A.~Kesting and M.~Treiber, ``Traffic flow dynamics: Data, models and
  simulation,'' 2013.

\bibitem{work2008ensemble}
D.~B. Work, O.-P. Tossavainen, S.~Blandin, A.~M. Bayen, T.~Iwuchukwu, and
  K.~Tracton, ``An ensemble kalman filtering approach to highway traffic
  estimation using gps enabled mobile devices,'' in \emph{Decision and Control,
  2008. CDC 2008. 47th IEEE Conference on}.\hskip 1em plus 0.5em minus
  0.4em\relax IEEE, 2008, pp. 5062--5068.

\bibitem{oh2002section}
J.-S. Oh, R.~Jayakrishnan, and W.~Recker, ``Section travel time estimation from
  point detection data,'' \emph{Center for Traffic Simulation Studies}, 2002.

\bibitem{zhan2013urban}
X.~Zhan, S.~Hasan, S.~V. Ukkusuri, and C.~Kamga, ``Urban link travel time
  estimation using large-scale taxi data with partial information,''
  \emph{Transportation Research Part C: Emerging Technologies}, vol.~33, pp.
  37--49, 2013.

\bibitem{jia2001pems}
Z.~Jia, C.~Chen, B.~Coifman, and P.~Varaiya, ``The pems algorithms for
  accurate, real-time estimates of g-factors and speeds from single-loop
  detectors,'' in \emph{Intelligent Transportation Systems, 2001. Proceedings.
  2001 IEEE}.\hskip 1em plus 0.5em minus 0.4em\relax IEEE, 2001, pp. 536--541.

\bibitem{de2008traffic}
C.~De~Fabritiis, R.~Ragona, and G.~Valenti, ``Traffic estimation and prediction
  based on real time floating car data,'' in \emph{Intelligent Transportation
  Systems, 2008. ITSC 2008. 11th International IEEE Conference on}.\hskip 1em
  plus 0.5em minus 0.4em\relax IEEE, 2008, pp. 197--203.

\bibitem{li2015inferring}
M.~Li, A.~Ahmed, and A.~J. Smola, ``Inferring movement trajectories from gps
  snippets,'' in \emph{Proceedings of the Eighth ACM International Conference
  on Web Search and Data Mining}.\hskip 1em plus 0.5em minus 0.4em\relax ACM,
  2015, pp. 325--334.

\bibitem{hofleitner2012learning}
A.~Hofleitner, R.~Herring, P.~Abbeel, and A.~Bayen, ``Learning the dynamics of
  arterial traffic from probe data using a dynamic bayesian network,''
  \emph{IEEE Transactions on Intelligent Transportation Systems}, vol.~13,
  no.~4, pp. 1679--1693, 2012.

\bibitem{morgul2013commercial}
E.~F. Morgul, K.~Ozbay, S.~Iyer, and J.~Holguin-Veras, ``Commercial vehicle
  travel time estimation in urban networks using gps data from multiple
  sources,'' in \emph{Transportation Research Board 92nd Annual Meeting}, no.
  13-4439, 2013.

\bibitem{gonzalez2007adaptive}
H.~Gonzalez, J.~Han, X.~Li, M.~Myslinska, and J.~P. Sondag, ``Adaptive fastest
  path computation on a road network: a traffic mining approach,'' in
  \emph{Proceedings of the 33rd international conference on Very large data
  bases}.\hskip 1em plus 0.5em minus 0.4em\relax VLDB Endowment, 2007, pp.
  794--805.

\bibitem{yuan2010t}
J.~Yuan, Y.~Zheng, C.~Zhang, W.~Xie, X.~Xie, G.~Sun, and Y.~Huang, ``T-drive:
  driving directions based on taxi trajectories,'' in \emph{Proceedings of the
  18th SIGSPATIAL International conference on advances in geographic
  information systems}.\hskip 1em plus 0.5em minus 0.4em\relax ACM, 2010, pp.
  99--108.

\bibitem{wu2004travel}
C.-H. Wu, J.-M. Ho, and D.-T. Lee, ``Travel-time prediction with support vector
  regression,'' \emph{IEEE transactions on intelligent transportation systems},
  vol.~5, no.~4, pp. 276--281, 2004.

\bibitem{yazici2014highway}
M.~Yazici, C.~Kamga, and K.~Ozbay, ``Highway versus urban roads: Analysis of
  travel time and variability patterns based on facility type,''
  \emph{Transportation Research Record: Journal of the Transportation Research
  Board}, no. 2442, pp. 53--61, 2014.

\bibitem{schmidhuber2015deep}
J.~Schmidhuber, ``Deep learning in neural networks: An overview,'' \emph{Neural
  networks}, vol.~61, pp. 85--117, 2015.

\bibitem{wang2016simple}
H.~Wang, Y.-H. Kuo, D.~Kifer, and Z.~Li, ``A simple baseline for travel time
  estimation using large-scale trip data,'' in \emph{Proceedings of the 24th
  ACM SIGSPATIAL International Conference on Advances in Geographic Information
  Systems}.\hskip 1em plus 0.5em minus 0.4em\relax ACM, 2016, p.~61.

\bibitem{grewal2007global}
M.~S. Grewal, L.~R. Weill, and A.~P. Andrews, \emph{Global positioning systems,
  inertial navigation, and integration}.\hskip 1em plus 0.5em minus 0.4em\relax
  John Wiley \& Sons, 2007.

\bibitem{lecun1995learning}
Y.~LeCun, L.~Jackel, L.~Bottou, C.~Cortes, J.~S. Denker, H.~Drucker, I.~Guyon,
  U.~Muller, E.~Sackinger, P.~Simard \emph{et~al.}, ``Learning algorithms for
  classification: A comparison on handwritten digit recognition,'' \emph{Neural
  networks: the statistical mechanics perspective}, vol. 261, p. 276, 1995.

\bibitem{cirecsan2012deep}
D.~C. Cire{\c{s}}an, U.~Meier, L.~M. Gambardella, and J.~Schmidhuber, ``Deep
  big multilayer perceptrons for digit recognition,'' in \emph{Neural networks:
  tricks of the trade}.\hskip 1em plus 0.5em minus 0.4em\relax Springer, 2012,
  pp. 581--598.

\bibitem{west2000neural}
D.~West, ``Neural network credit scoring models,'' \emph{Computers \&
  Operations Research}, vol.~27, no.~11, pp. 1131--1152, 2000.

\bibitem{hinton2006reducing}
G.~E. Hinton and R.~R. Salakhutdinov, ``Reducing the dimensionality of data
  with neural networks,'' \emph{science}, vol. 313, no. 5786, pp. 504--507,
  2006.

\bibitem{hornik1989multilayer}
K.~Hornik, M.~Stinchcombe, and H.~White, ``Multilayer feedforward networks are
  universal approximators,'' \emph{Neural networks}, vol.~2, no.~5, pp.
  359--366, 1989.

\bibitem{haykin2009neural}
S.~S. Haykin, \emph{Neural networks and learning machines}.\hskip 1em plus
  0.5em minus 0.4em\relax Pearson Upper Saddle River, NJ, USA:, 2009, vol.~3.

\bibitem{hecht1988theory}
R.~Hecht-Nielsen \emph{et~al.}, ``Theory of the backpropagation neural
  network.'' \emph{Neural Networks}, vol.~1, no. Supplement-1, pp. 445--448,
  1988.

\bibitem{NYCURL}
``Flicker nyc coordinates,'' \url{https://www.flickr.com/places/info/2459115}.

\bibitem{wang2014travel}
Y.~Wang, Y.~Zheng, and Y.~Xue, ``Travel time estimation of a path using sparse
  trajectories,'' in \emph{Proceedings of the 20th ACM SIGKDD international
  conference on Knowledge discovery and data mining}.\hskip 1em plus 0.5em
  minus 0.4em\relax ACM, 2014, pp. 25--34.

\end{thebibliography}
\newpage
\end{document}